\documentclass[runningheads]{llncs}
\usepackage{graphicx}

\usepackage[utf8]{inputenc} 
\usepackage[T1]{fontenc}    
\usepackage{xcolor}         
\usepackage{hyperref}       

\usepackage{url}            
\usepackage{amsfonts,amsmath,amssymb} 
\usepackage{nicefrac}       
\usepackage{textcomp}       
\usepackage{adjustbox}      
\usepackage{booktabs, multirow, tabularx, array}        
\usepackage{threeparttable}

\begin{document}
\title{Large Neural Networks Learning from Scratch with Very Few Data and without Explicit Regularization \thanks{The work of Christoph Linse was supported by the Bundesministerium f{\"u}r Wirtschaft und Klimaschutz through the Mittelstand-Digital Zentrum Schleswig-Holstein Project.}}
\titlerunning{Large Neural Networks Learning from Scratch with Very Few Data}
%
\author{Christoph Linse\orcidID{0000-0002-7039-5189} \and
Thomas Martinetz\orcidID{0000-0002-4539-4475}}
%
\authorrunning{C. Linse, T. Martinetz}
%
\institute{Institute for Neuro- and Bioinformatics - University of L{\"u}beck, Ratzeburger Allee 160, 23562 L{\"u}beck, Germany, \\
Email: \email{\{linse,martinetz\}@inb.uni-luebeck.de}
\\
\url{https://www.inb.uni-luebeck.de/en/home.html}}
\maketitle              


\begin{abstract}
    Recent findings have shown that highly over-parameterized Neural Networks generalize without pretraining or explicit regularization. It is achieved with zero training error, i.e., complete over-fitting by memorizing the training data. This is surprising, since it is completely against traditional machine learning wisdom. In our empirical study we fortify these findings in the domain of fine-grained image classification. We show that very large Convolutional Neural Networks with millions of weights do learn with only a handful of training samples and without image augmentation, explicit regularization or pretraining. We train the architectures ResNet018, ResNet101 and VGG19 on subsets of the difficult benchmark datasets Caltech101, CUB\_200\_2011, FGVCAircraft, Flowers102 and StanfordCars with 100 classes and more, perform a comprehensive comparative study and draw implications for the practical application of CNNs. Finally, we show that a randomly initialized VGG19 with 140 million weights learns to distinguish airplanes and motorbikes with up to 95\% accuracy using only 20 training samples per class.

\end{abstract}

\section{Introduction}

Large Deep Convolutional Neural Networks (CNNs) show excellent generalization capabilities in a variety of tasks, particularly in image recognition. Often, their performance can be boosted by increasing the depth or the width of the architectures, which is bound to an increase in trainable parameters.
The good generalization capabilities are often attributed to a suitable bias within modern CNN architectures and effective regularization strategies like weight decay or dropout. 

However, generalization is also present in highly over-paramete-rized regimes and without explicit regularization, where even random label assignments or random image data can be memorized \cite{zhang_understanding_2017}. Belkin called it the '"modern" interpolation regime' \cite{belkin_reconciling_2019}.
The generalization capabilities of large neural networks do not seem to be harmed by the millions of parameters in today's popular architectures. Due to their large, inherent capacity the training error is zero after training by Empirical Risk Minimization (ERM), i.e.\ the training data is memorized, but, nevertheless, the test error can be unexpectedly low, as shown in~\cite{zhang_understanding_2017}. This is in contrast to standard machine learning wisdom, where one would not expect any learning in these highly over-parameterized regimes. The data is highly over-fitted. Statistical learning theory based on uniform convergence requires a high probability that there is no bad solution with a large deviation between its empirical and its true risk, i.e. the training and the generalization error. This can only be expected if the number of training samples is (much) larger than the underlying VC-dimension of the classifier. However, this is not the case in over-parameterized regimes. In over-parameterized regimes bad solutions with zero training but large generalization error do exist. Interestingly, it seems that they hardly occur in general.  

In this paper, we empirically study the generalization of large, highly over-para\-meterized networks with a focus on fine-grained image classification. Usually, fine-grained classification is tackled using various techniques to boost the generalization abilities, such as image augmentation, regularization, novel loss functions, semi-supervised learning or pretrained networks. In this paper, we want to study the extreme case of very large networks with 10-140 million weights and just a few training samples of 10-67 per class, with no augmentation, no pretraining and no explicit regularization. We train models with random weight initialization and significantly more classes than samples per class, and choose complex, fine-grained classification problems. The generalization is measured for the popular, large models ResNet018, ResNet101 and VGG19 with increasing numbers of weights. We provide an extensive range of experiments and train hundreds of models on subsets of the Caltech101, CUB\_200\_2011, FGVCAircraft, Flowers102 and StanfordCars datasets.

This work is not related to transfer learning, which harnesses the well-known power of pretrained networks to boost generalization capabilities \cite{AlLiBaMa19}. Such models are pretrained on large-scale datasets like ImageNet \cite{deng2009imagenet} and are subsequently fine-tuned on a smaller dataset to solve a specific problem. However, during pretraining the total number of train images is similar to the number of trainable parameters, which is far from the highly over-parameterized regime that we study in this paper. Here, we use completely randomly initialized networks.

The paper is structured as follows. Section~\ref{sec:materials_and_methods} provides all details about our experimental setup including the fine-grained classification datasets and the architectures.
The results are presented and discussed in Section~\ref{sec:results_and_discussion}.
The paper completes with our conclusions in Section~\ref{sec:conclusion}.

\section{Materials and Methods}
\label{sec:materials_and_methods}

\subsection{Datasets}
\label{sec:materials_and_methods:datasets}

\begin{figure}[t]
	\centering
	\includegraphics[width=0.8\linewidth]{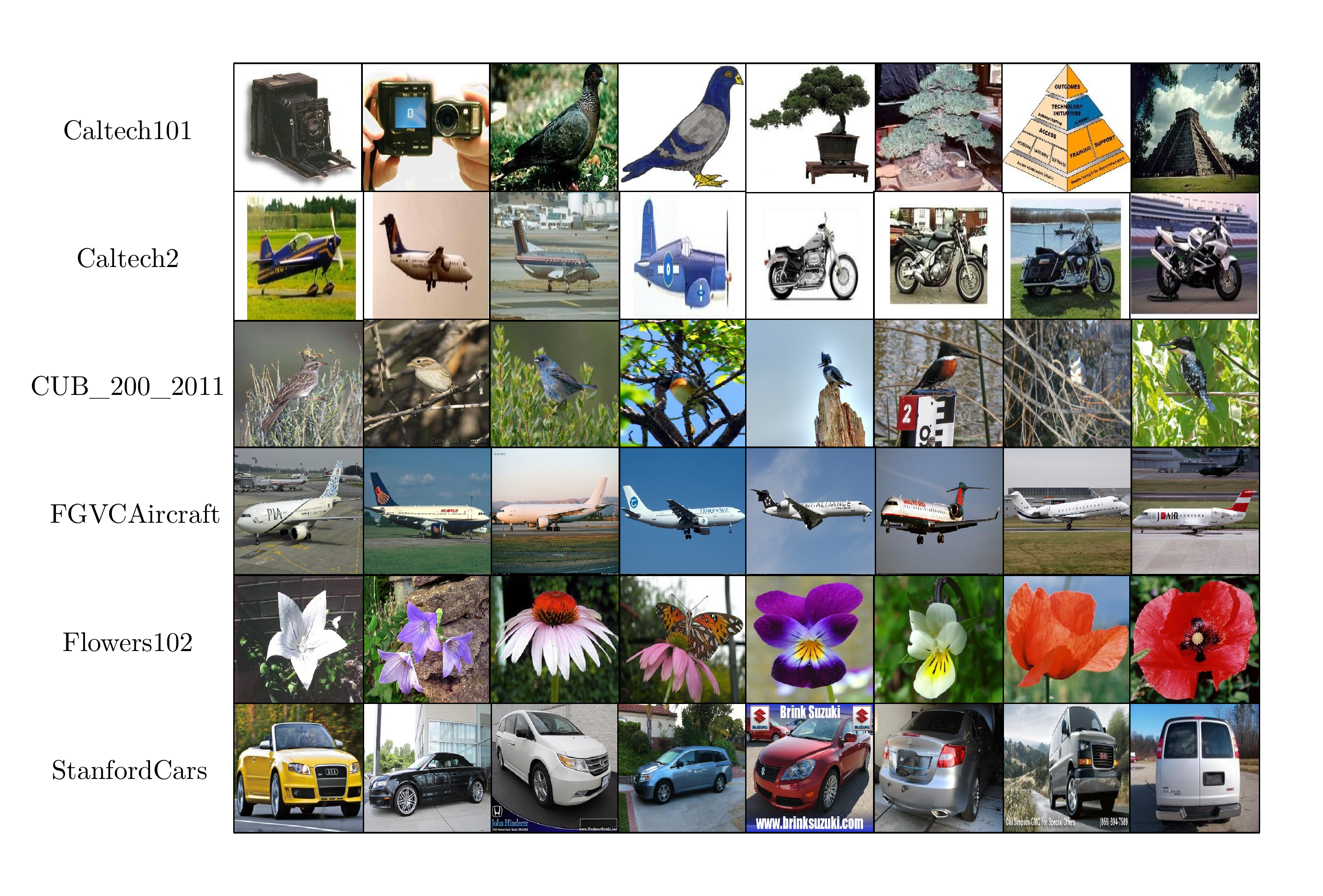}
	\caption{Fine-grained image classification datasets. Two examples of four different classes are shown, respectively. For Caltech2 with its two classes four examples per class are depicted.}
    \label{fig:datasets_example_images}
\end{figure}

The fine-grained classification datasets are summarized in Table~\ref{tab:benchmark_datasets}, where the number of classes, the size of the datasets and statistics about the number of images per class are presented.
Figure~\ref{fig:datasets_example_images} provides example images from every dataset showing 2 images for 4 classes, respectively.

\begin{table}[b]
\caption{Detailed structure of the benchmark datasets.}
\label{tab:benchmark_datasets}
\centering
\begin{adjustbox}{width=\linewidth}

\begingroup
\setlength{\tabcolsep}{8pt} 
\begin{tabular}{ccccccc}
& & & & & & \\

                                &       &                      & \multicolumn{4}{|c|}{\textbf{Number of images in/of...}}                         \\
\textbf{Dataset}                & \textbf{Mode} & \textbf{Classes} & \multicolumn{1}{|c}{\textbf{dataset}} & \textbf{smallest class} & \textbf{largest class} & \multicolumn{1}{c|}{\textbf{avg. per class}} \\
\midrule
Caltech101                      & -     & 101                  & 8677    & 31             & 800           & 85.91             \\
Caltech2                        & -     & 2                    & 1596    & 798            & 798           & 798.00               \\
\midrule
\multirow{2}{*}{CUB\_200\_2011} & train & \multirow{2}{*}{200} & 5994    & 29             & 30            & 29.97             \\
                                & test  &                      & 5794    & 11             & 30            & 28.97             \\
\midrule
\multirow{2}{*}{FGVCAircraft}  & train & \multirow{2}{*}{100} & 6667    & 66             & 67            & 66.67             \\
                                & test  &                      & 3333    & 33             & 34            & 33.33             \\
\midrule
\multirow{2}{*}{Flowers102}     & train & \multirow{2}{*}{102} & 2040    & 20             & 20            & 20.00               \\
                                & test  &                      & 6149    & 20             & 238           & 60.28               \\
\midrule
\multirow{2}{*}{Stanford Cars}  & train & \multirow{2}{*}{196} & 8144    & 24             & 68            & 41.55             \\
                                & test  &                      & 8041    & 24             & 68            & 41.03             \\

\end{tabular}
\endgroup
\end{adjustbox}
\end{table}

\paragraph{Caltech101} \cite{li_fei_fei_learning_2004} was published in 2004. It contains 101 categories and a background class, which we ignore in our experiments. The categories offer a wide degree of diversity including objects like laptops, smartphones and motorbikes, but also animal species and plants. 8677 images were collected manually from the internet and scaled to have a width of about 300 pixels. Sometimes, the objects are shown in a cluttered environment and sometimes they are seen in front of a white background.

\paragraph{Caltech2} is a two-class subset of Caltech101 with 798 images showing motorbikes and 798 images showing airplanes.

\paragraph{CUB\_200\_2011} is the Caltech-UCSD Birds-200-2011 dataset \cite{wah_caltech_ucsd_2011} from 2011, which extends the Caltech-UCSD Birds 200 dataset \cite{welinder_caltech_ucsd_2010}. It contains a total of 5994 train and 5794 test images with a mean width of about 470 pixels. There are 200 different bird categories and each category is represented with about 30 training images. Variations in color of the plumage, lighting, perspectives and poses introduce a large intra-class variance while the differences between bird categories can be subtle, which makes this data set particularly challenging.

\paragraph{FGVCAircraft} or the Fine-Grained Visual Classification of Aircraft dataset~\cite{maji_fine_grained_2013} was published in 2013 and contains 6667 training  and 3333 test images showing 100 different airplane models. The images typically have wide aspect ratios and a width of about 1100 pixels. The data is split into an official training and test set with about 67 and 33 images per class, respectively. 
Airplanes are rigid objects and therefore do not pose the challenge of deformation. However, there is a notable degree of intra-class variance in the visual appearance due to advertisement, airlines and perspective.

\paragraph{Flowers102} was published in 2008 as the 102 Category Flower dataset~\cite{nilsback_automated_2008}. In average the images are 630 pixels wide and show one or several blossom instances. In order to be consistent with the other datasets, which have training and test sets only, we train with the combination of the training and validation set of Flowers102, leading to 20 training images for each flower type and 2040 images in total. For testing there are 6149 images. While flowers can have small inter-class variations, visual properties like scale, pose, light and also flower subtypes (e.g. color variants) can lead to strong intra-class variation. Therefore, Flowers102 is considered to be a challenging dataset. 

\paragraph{StanfordCars} \cite{krause_3d_2013} from Stanford University was presented in 2013 and is composed of 196 different car models. The 8144 training and 8041 test images have an average width of 700 pixels. Typically, only one car is shown at once. The imaging conditions and the locations of the cars are not restricted to a single environment. Sometimes, a car is on a road, other images have been taken indoors and with a stylized background. Also, the perspective is not constrained, such that the image recognition system has to learn features from the front, side and rear of a car model.

\subsection{Network Architectures}
\label{sec:materials_and_methods:architectures}

ResNet018~\cite{he_deep_2016} is employed as a baseline model. It is moderately sized with 11.2 million weights, but still in the over-parameterized regime with regard to the small-scale datasets in the study. 
Then, we use ResNet101 as a much deeper model with 101 layers and almost four times as many weights as ResNet018. 
Third, we use the VGG19~\cite{simonyan_very_2015} architecture with a total of 140.0 million weights. A trained VGG19 model needs more than half a gigabyte when saved on the disk. The VGG19 consists of a set of 16 convolutional layers and 3 fully connected ones. The majority of the weights are located at the fully connected part of the network.
More detailed information about the architectures is provided in Table~\ref{tab:model_size}, where the number of parameters, the number of numerical operations, the model size and the memory consumption on the GPU when performing a forward pass are summarized.

\begin{table}[tb]
\caption{Details on the architectures. The memory consumption during the forward pass was measured using a batch size of 32 and an image size of $3 \times 224 \times 224$ pixels.}
\label{tab:model_size}
\centering
\begin{adjustbox}{width=1\linewidth}
\begin{tabular}{ccccc}
Model     & Parameters($10^6$) & Mult-Adds($10^9$) & Size(MB) & GPU Memory Forward Pass(MB) \\
\midrule
ResNet018 & 11.2             & 1.8            & 43.9      & 1825                         \\
ResNet101 & 42.7             & 7.8            & 167.5     & 5119                         \\
VGG19     & 140.0              & 19.6           & 546.9     & 6009                         \\
\end{tabular}
\end{adjustbox}
\end{table}

\subsection{Training Set-up}
\label{sec:materials_and_methods:training}
The network weights are initialized using Kaiming initialization~\cite{he_delving_2015}. Training minimizes the cross-entropy loss for 300 epochs. At the end of the training procedure the classification error on the training set is always zero. The Lamb optimizer~\cite{you2020large} is used to get a stable training process with smooth learning curves. The automatic differentiation framework of Pytorch is used~\cite{paszke2017automatic}. 
The initial learning rate of 0.0015 is decreased until $1.5 \cdot 10^{-6}$ in a step-wise manner. The batch size is 32.
No image augmentation is performed. Also, there is no explicit regularization like weight decay or dropout. 
In our data pre-processing pipeline the input images are reshaped to have $224 \times 224$ pixels. In a normalization step the mean RGB value of the entire dataset is subtracted from each image and we divide by the standard deviation.
5 models are trained in each setting to enable a statistical validation of their performance metrics.
Depending on the dataset size, it takes between 30 minutes and a few hours to train a single model on a single NVIDIA GTX 2080 Ti graphics card.

When dealing with very small datasets (between 40 and 8144 train samples) statistical fluctuations have to be taken into account. We address this issue by presenting results on 5 different datasets to enable the comparison between different data distributions and to draw meaningful conclusions. For each single experiment we train and test 5 networks with different seeds. Please note, that we also choose different data subsets for the 5 networks when training on the Caltech dataset.

\section{Results and Discussion}
\label{sec:results_and_discussion}

The models are trained on different subsets with increasing number of training samples per class. Table~\ref{tab:accuracies_benchmark_datasets} and Table~\ref{tab:accuracies_caltech101} show the obtained accuracies on the test sets. For the Caltech101 dataset we have a separate table, since there is no official split in training and test set and we experimented with different splits.

\begin{table}[tb]
\caption{Test accuracy of different classifiers on the benchmark datasets with various number of training samples per class.
For each configuration 5 models were trained with different seeds and initializations. The results show the mean test performance and its standard deviation.}
\label{tab:accuracies_benchmark_datasets}
\centering
\begin{adjustbox}{width=0.7\linewidth}
\begin{tabular}{cccccc}
& & & & \\

\textbf{Dataset} & \textbf{Classes} & \textbf{Model} & \textbf{Samples p. class} & \textbf{Accuracy} & \textbf{Std} \\
\midrule
\multirow{7}{*}{CUB\_200\_2011} & \multirow{7}{*}{200} & ResNet018 & 20       & 15.04 & 0.60 \\
 & & ResNet101 & 20       & \textbf{22.45} & \textbf{0.58} \\
 & & VGG19     & 20       & 16.99 & 0.80 \\
 & & ResNet018 & 30 (all) & 22.66 & 0.75 \\
 & & ResNet101 & 30 (all) & \textbf{29.96} & \textbf{1.03} \\
 & & VGG19     & 30 (all) & 24.99 & 0.97 \\
 & & SVM (HOG + RGB) \cite{alter2017exploration} & 30 (all) & 5.00 & - \\

\midrule
\multirow{7}{*}{FGVCAircraft}   & \multirow{7}{*}{100} & ResNet018 & 20       & 16.03 & 1.09 \\
 & & ResNet101 & 20       & 25.01 & 0.46 \\
 & & VGG19     & 20       & \textbf{36.63} & \textbf{0.85} \\
 & & ResNet018 & 67 (all) & 49.88 & 0.59 \\
 & & ResNet101 & 67 (all) & 58.91 & 0.24 \\
 & & VGG19     & 67 (all) & \textbf{68.78} & \textbf{0.26} \\
 & & SVM, Spatial Pyramid \cite{maji_fine_grained_2013}     & 67 (all) & 48.69 & - \\
\midrule
\multirow{7}{*}{Flowers102}     & \multirow{7}{*}{102} & ResNet018 & 10       & 37.21 & 0.82 \\
 & & ResNet101 & 10       & \textbf{40.45} & \textbf{2.07} \\
 & & VGG19     & 10       & 36.99 & 1.06 \\
 & & ResNet018 & 20 (all) & \textbf{54.66} & \textbf{1.52} \\
 & & ResNet101 & 20 (all) & \textbf{54.87} & \textbf{3.40} \\
 & & VGG19     & 20 (all) & 53.05 & 0.76 \\
 & & SVM (HSV) \cite{nilsback_automated_2008}       & 10 & 43.0 & - \\
\midrule
\multirow{7}{*}{StanfordCars}   & \multirow{7}{*}{196} & ResNet018 & 20       & 5.50  & 0.28 \\
 & & ResNet101 & 20       & \textbf{6.79}  & \textbf{0.53} \\
 & & VGG19     & 20       & 4.72  & 1.09 \\
 & & ResNet018 & $\approx 42$ (24-68 p.cl.) & 20.33 & 0.66 \\
 & & ResNet101 & $\approx 42$ (24-68 p.cl.) & \textbf{24.34} & \textbf{1.27} \\
 & & VGG19     & $\approx 42$ (24-68 p.cl.) & 16.14 & 4.75 \\
 & & SVM, Spatial Pyramid \cite{krause_3d_2013} & $\approx 42$ (24-68 p.cl.) & 69.5 & - \\

\end{tabular}
\end{adjustbox}
\end{table}

\begin{table}[tb]
\caption{Performance of ResNet018, ResNet101 and VGG19 on the Caltech101 dataset for different splits into training and test set. The results show the mean performance and standard deviation of 5 trained models.}
\label{tab:accuracies_caltech101}
\centering
\begin{adjustbox}{width=0.7\linewidth}
\begin{threeparttable}[b]

\begin{tabular}{ccccc}
& & & & \\

\textbf{Model} & \textbf{Train samples} & \textbf{Test samples} & \textbf{Accuracy} & \textbf{Std} \\
\midrule

ResNet018 & \multirow{3}{*}{10 per class} & \multirow{3}{*}{20 per class}     & 27.24 & 1.12 \\
ResNet101 & & & 32.79 & 1.28 \\
VGG19     & & & \textbf{39.41} & \textbf{0.57} \\
\midrule
ResNet018 & \multirow{3}{*}{20 per class} & \multirow{3}{*}{10 per class}     & 42.30 & 0.98 \\
ResNet101 & & & 43.96 & 1.73 \\
VGG19     & & & \textbf{54.38} & \textbf{1.01} \\
\midrule
ResNet018 & \multirow{3}{*}{80.00\% (25-640 p.cl.)~} & \multirow{3}{*}{~20.00\% (6-160 p.cl.)} & 65.90 & 0.52 \\
ResNet101 & & & 65.27 & 1.53 \\
VGG19     & & & \textbf{71.11} & \textbf{0.83} \\
\midrule
ResNet018 & \multirow{3}{*}{rest (21-790 p.cl.)}    & \multirow{3}{*}{10 per class}      & 66.08 & 1.24 \\
ResNet101 & &  & 66.32 & 1.87 \\
VGG19     & & & \textbf{71.52} & \textbf{0.59} \\
\midrule
DAGSVM \cite{zhang_SVM_KNN_2006} & 15 per class      & rest (16-785 p.cl.)     & 59.05\tnote{1} & 0.56 \\
DAGSVM \cite{zhang_SVM_KNN_2006} & 30 per class      & rest (1-770 p.cl.)     & 66.23\tnote{1} & 0.48 \\
\bottomrule
\end{tabular}
\begin{tablenotes}
    \item [1] Mean recognition rate per class.
\end{tablenotes}
\end{threeparttable}
\end{adjustbox}
\end{table}

\begin{figure}[tb]
	\centering
	\includegraphics[width=0.6\linewidth]{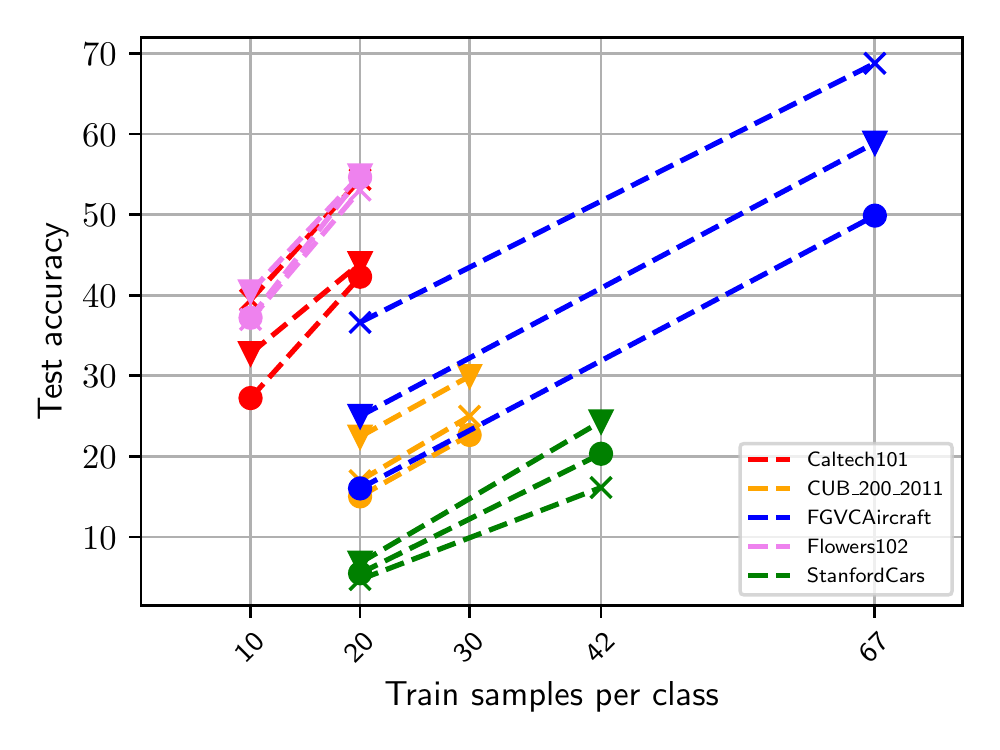}
	\caption{Performance of ResNet018, ResNet101 and VGG19 on different splits of fine-grained image classification datasets. ResNet018 is denoted as circles, ResNet101 as triangles and VGG19 as x. The colors correspond to the different datasets. The results show the mean performance of 5 trained models.}
    \label{fig:accuracies_datasets}
\end{figure}

In Figure~\ref{fig:accuracies_datasets} the results are illustrated more clearly. All networks with their 10-140 million parameters do learn with a handful of training data per class. Before training, their accuracy is 0.01-0.005 (random guessing), and after training the accuracy has increased by a factor of 10-70. It is astonishing, that a classifier with 140 million parameters (VGG19) learns to discriminate 100 different aircraft models with only 67 example images per class up to 70\% accuracy, which is an increase by a factor of 70. It is correct in more than 2 of 3 cases in a task which is definitely difficult also for humans. This is in great contrast to what traditional machine learning wisdom would have expected in these highly over-parameterized regimes. 
The Flowers102 dataset provides similar results. With only 20 example images per class the classifier is able to learn to discriminate 102 flowers up to an accuracy of more than 50\%. Instead of being correct in every 100th case (before training), after training it is correct in every second case. The increase in accuracy is by a factor of 50. For the 200 birds ResNet101 achieves the best results with an increase in accuracy by a factor of 60 by training with 30 images per class. The variance is always below 5\%. For these large networks there exist bad zero training error solutions, even solutions with zero test accuracy. However, these bad solutions did not occur. The accuracies for different seeds hardly vary.   

On the datasets we studied, the most successful model is ResNet101, which is also the deepest one in this study. It is the best generalizing model in 6 of 8 cases. VGG19, the model with the highest number of weights, is the best generalizing model in 2 of 8 cases.
The model with the fewest parameters, ResNet018, has the worst results on the fine-grained image classification tasks, even though it should generalize better when trained on small datasets according to traditional machine learning wisdom. 

Of course, these results are not state-of-the-art but can be further improved by standard tricks like data augmentation, regularization or pretrained weights. Instead, we demonstrate that learning takes place easily far outside the orbits of traditional machine learning wisdom, often just as good or even better than it was state-of-the-art a few years ago within the orbits of traditional thinking. For this purpose, we have added some state-of-the-art results from then in the Tables~\ref{tab:accuracies_benchmark_datasets} and~\ref{tab:accuracies_caltech101}, usually achieved with Support-Vector Machines on elaborated handcrafted features. 

\begin{figure}[tb]
	\centering
	\includegraphics[width=0.6\linewidth]{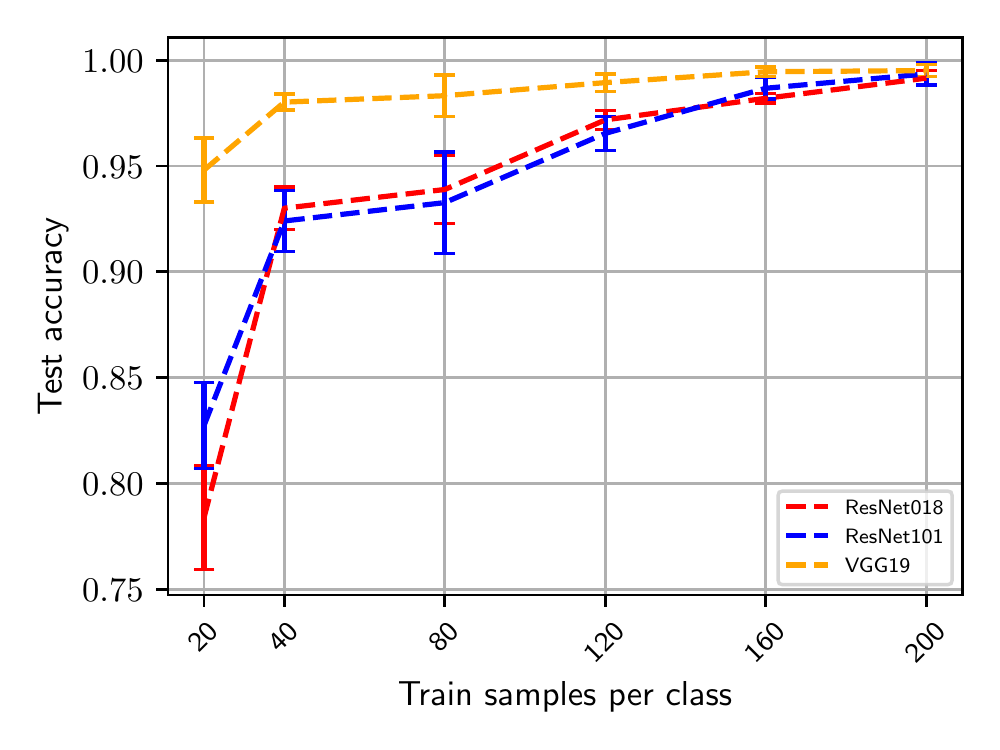}
	\caption{Test accuracy on different splits of the Caltech2 dataset consisting of images showing airplanes and motorbikes. The results show the mean performance and standard deviation of 5 models.}
    \label{fig:accuracies_caltech2}
\end{figure}

Another interesting observation in Figure~\ref{fig:accuracies_datasets} is that the accuracy curves of different architectures plotted against the number of training samples per class have no intersections. This means that the performance ranking of the architectures is consistent over the number of training samples. 
For CUB\_200\_2011, Flowers102 and StanfordCars the best generalizing model is ResNet101, disregarding the size of the training set. For FGVCAircraft the VGG19 net consistently provides the best results.
Thus, the experiments suggest that the selection of the best model depends on the dataset and not on the number of training images. In other words, the data distribution chooses the architecture, not the size of the training dataset. Of course, it is not clear to what extend this relationship can be generalized. 

\subsection{Further Experiments on the Caltech101 Dataset}

In order to substantiate the findings further experiments are performed on the Caltech101 dataset.
Table~\ref{tab:accuracies_caltech101} presents the accuracy of ResNet018, ResNet101 and VGG19 using different splits of the Caltech101 dataset, which lacks an official split. Interestingly, the network with the highest number of parameters, VGG19, has always the highest generalization ability with a maximum of $71.52(59)\%$ test accuracy. Again, the relative order of the models in terms of generalization does not depend on the choice of the training split.

This finding is studied more closely by ablating the Caltech101 dataset. We built a subset of Caltech101 by selecting 798 motorbike and 798 aircraft images as they are the classes with the largest number of samples. We trained the three architectures on this binary classification problem using different training and test splits and kept the previous hyperparameters. 
Figure~\ref{fig:accuracies_caltech2} shows the performances of 5 models trained on each configuration. The test accuracy is plotted against the number of training images. Again, the models always have zero training error. 

Just as in the case of the complete Caltech101 dataset, VGG is the best generalizing model. It achieves a test accuracy of about 95\% with only 20 training samples per class, which is highly unexpected in such a highly over-parameterized regime.
The curves of ResNet018 and ResNet101 are very close and cannot be distinguished from each other. Only for 20 training images per class ResNet101 has an about 5\% higher test accuracy than ResNet018. Again, the architecture with more weights performs better in this regime, and again the accuracy curves of single architectures with respect to the number training  samples do not intersect, which is in line with our previous observations. Already for 200 training samples all three models converge to an almost perfect classification accuracy of 99\%. The variance is below 1\%. Again, bad solutions, which exist, do not occur.

\section{Conclusion and Outlook}
\label{sec:conclusion}

We studied the generalization abilities of deep, high capacity Convolutional Neural Networks in the context of fine-grained image classification with very small training datasets. We trained the frequently used CNNs ResNet018, ResNet101 and VGG19 without augmentation, explicit regularization or pretrained weights in order to measure their genuine generalization ability in highly over-parameterized regimes on the challenging benchmark datasets Caltech101, CUB\_200\_2011, FGVCAircraft, Flowers102 and StanfordCars. We were able to show that the generalization error is remarkably low already with very few training data, and that the two larger CNNs performed consistently better than the smallest one. For a binary classification problem we could show that a network with 140 million weights (VGG19) reaches 95\% accuracy with only 20 training samples per class. This is against the narrative that Deep Neural Networks require large training data sets. We observed, that certain architectures work better on certain datasets, no matter the size of the training split. The relative generalization abilities between architectures do not seem to depend on the number of training images. This supports the idea that problems prefer certain models, but models do not prefer a specific training set size more than other models. 
We conclude, that very large Neural Networks can learn with very few data, also without any data augmentation, explicit regularization or pretraining.  Performance even can benefit from an increased number of network weights in these highly over-paramete-rized regimes. Training up to zero training error, i.e., complete over-fitting by memorizing the training data, does not harm generalization. All this is against traditional machine learning wisdom. At least in case of binary image classification problems, only a few hundred training samples can be sufficient also for very large CNNs to provide best possible solutions. In principle, bad solutions do exist in these highly over-parameterized regimes, but they did not occur.  

Our empirical findings are restricted to CNNs and image classification tasks. We expect, that these findings extrapolate to other task domains and network architectures, but this remains to be shown. The observation, that the accuracy curves of different architectures do not intersect could be validated for large scale datasets in future work. Moreover, the study could be enriched with more recent architectures of different sizes or networks that share the same basic architecture and only differ in the number of filters inside the layers.

We need a new understanding of learning, which extends classical statistical learning theory into over-parameterized regimes. 
This will require abandoning or at least expanding the concept of uniform convergence and traditional capacity measures of classifiers. The usual narratives of traditional machine learning wisdom of "over-fitting" and "the larger the network the more data is needed" do not apply in this realm.

%
%
%
%
\bibliographystyle{splncs04} 
\bibliography{References}
\end{document}